\documentclass[letterpaper, 10 pt, conference]{ieeeconf}  

\IEEEoverridecommandlockouts                              

\usepackage{graphics} 
\usepackage{epsfig} 
\usepackage{float}
\usepackage[caption=false,font=footnotesize]{subfig}

\title{\LARGE \bf

A behavioural transformer for effective collaboration between a robot and a non-stationary human
}

\author{Ruaridh Mon-Williams$^{1}$, Theodoros Stouraitis$^{2}$ and Sethu Vijayakumar$^{1,3}$
\thanks{*This work was supported by the Edinburgh Centre for Robotics.}
\thanks{$^{1}$Ruaridh Mon-Williams and Sethu Vijayakumar are with the School of Informatics, University of Edinburgh, Edinburgh EH8 9AB, U.K. (e-mail:
ruaridh.mw@ed.ac.uk, sethu.vijayakumar@ed.ac.uk)}%
\thanks{$^{2}$Theodoros Stouraitis is with the Honda Research
Institute Europe (HRI-EU), 63073 Offenbach am Main, Germany (e-mail:
theostou@honda-ri.de).%
}
\thanks{$^{3}$Sethu Vijayakumar is also with The Alan Turing Institute, London, U.K.}}

\usepackage{amsmath}
\usepackage{amsfonts}
\usepackage{graphicx}
\usepackage{float} 
\usepackage{cleveref}

\usepackage{algpseudocode}
\usepackage{algorithm}
\usepackage[style=ieee]{biblatex}
\usepackage{xcolor}

\begin{document}

\maketitle
\thispagestyle{empty}
\pagestyle{empty}

\begin{abstract}
A key challenge in human-robot collaboration is the non-stationarity created by humans due to changes in their behaviour. This alters environmental transitions and hinders human-robot collaboration. We propose a principled meta-learning framework to explore how robots could better predict human behaviour, and thereby deal with issues of non-stationarity. On the basis of this framework, we developed Behaviour-Transform (BeTrans). BeTrans is a conditional transformer that enables a robot agent to adapt quickly to new human agents with non-stationary behaviours, due to its notable performance with sequential data. We trained BeTrans on simulated human agents with different systematic biases in collaborative settings. We used an original customisable environment to show that BeTrans effectively collaborates with simulated human agents and adapts faster to non-stationary simulated human agents than SOTA techniques.
\end{abstract}

\section{INTRODUCTION}

Reinforcement learning (RL) provides a powerful framework for robots to learn policies, but such learning has focused on single-agent tasks and few robots can interact effectively with humans \cite{xie_learning_2020, nguyen_review_2019, stouraitis_online_2020}.
In contrast to existing robotic agents, humans are proficient in collaborating with other humans. Yet, our understanding of the joint action mechanisms is still the subject of research~\cite{obhi_moving_2011}. A key aspect of joint action, identified by Sheridan~\cite{sheridan_eight_nodate}, is to acquaint both human and robot agents with models of their partners so that  predicting the behaviour of the partner agent is possible. 

RL in a multi-agent setting requires the robot agent to account for the human's behaviour \cite{kim_influencing_2022}. The difficulty is that human behaviour is typically non-stationary and shows individual differences (requiring ad-hoc coordination). This means that robot agents need to predict human actions based on the observation history, but this has proved extremely challenging to date \cite{xie_learning_2020, yang_optimal_2022}. In the current work, we focus on how a robot agent could predict the behaviour of a collaborative human agent with non-stationary behaviour.

To understand such a challenge better, let us consider two players passing a ball between themselves. The speed of the ball combined with the perceptual and motor lags of the humans means that each player must predict where the other player will hit the ball~\cite{bobu_less_2020}. The human's behavior is affected by long and short-term historical dependencies, such as the pose of the partner, previous hit, \textit{etc}. To accurately predict the goal of the next hit by the human, both short term (\textit{e.g.} previous partner pose) and long term (\textit{e.g.} direction of motion of the partner) information should be used~\cite{nomura_prediction_2008}. To effectively deal with this dynamic environment, it is imperative for each player to update their predictions and their strategies according to their partner's behaviour~\cite{nikolaidis_human-robot_2017}. 


To study such a cooperative problem, we created a 'virtual' ball hitting game to explore the effectiveness of a meta-learning approach. Our approach uses a transformer to learn a latent variable that allows the robot agent to dynamically update its behaviour. We explored whether this could help a robot agent predict where a simulated human agent would hit the ball and thereby collaborate successfully (despite non-stationarity). This manuscript details the following contributions that resulted from these endeavours:

\begin{itemize}
  \item Formalisation of the 'human-robot cooperative RL problem with non-stationary human agent behaviours' as a zero-shot meta-learning theoretic framework.
  \item Development of a transformer-based RL approach to; (i) infer the latent state of a human agent whose behaviour changes over time, and (ii) enable a robot agent to effectively collaborate  with a human agent. 
  \item Demonstration that the behavioural transformer approach - BeTrans - can effectively cooperate with simulated human agents and outperform state-of-the-art methods, evaluated in a novel custom environment.
\end{itemize}

Next, we summarize the relevant related work and highlight how this differs to our proposed approach.


\subsection{Related work}


\subsubsection*{Non-Stationarity in RL}

Single-agent RL facilitates optimal decision-making in static environments. In human-robot collaboration scenarios, the human is typically considered part of the environment. This makes the environment (transition function) time-dependent, due to the non-stationary human behavior. As a result, traditional RL approaches produce sub-optimal results in such scenarios~\cite{papoudakis_dealing_nodate, xie_learning_2020}. This challenge is the focus of the current work.

\subsubsection*{Multiagent RL}
Multiagent reinforcement learning (MARL) aims to tackle non-stationarity~\cite{albrecht_autonomous_2018}, but it typically uses a centralised framework. This approach is not suitable for human-robot collaboration settings as it is not possible to control both agents. In current work, the human agent is treated as part of the environment and modelled separately. 

\subsubsection*{Learning Latent Strategies}


One way to model a human agent is by assuming that their behaviour is governed by a latent state. This approach assumes there is an underlying structure in the human policy space with a low dimensionality (\textit{e.g.} high-level human goals) that can capture the low-level policies executed by a human.
Robots can differentiate between different human behaviors by treating the latent variables as distinct tasks ~\cite{wang_uencing_nodate, parekh_rili_2022, xie_learning_2020}. This paradigm approach is also followed in our work. 

LILI, SILI, and RILI \cite{wang_uencing_nodate, parekh_rili_2022, xie_learning_2020} learn high-level latent strategies, utilising  feed-forward (MLP) or recurrent (RNN) neural networks in encoder-decoder blocks, and condition single-agent RL policies on them.
However, MLPs struggle with sequential modeling due to their lack of memory and an inability to handle variable-length sequences. On the other hand, RNNs are effective in capturing short-term dependencies, but struggle with modeling long-term dependencies in sequential data~\cite{karita_comparative_2019}. These state-of-the-art approaches acknowledge that the learnt policies can be too brittle to handle noisy interactions with a human, and may struggle with long-term dependencies and behaviours that vary within episodes. Our current approach utilises transformers to address the issues highlighted above.

\subsubsection*{Transformers}
These deep learning models are adept at processing sequential data and are proficient in discerning semantic relationships and capturing the long-term dependencies that cause difficulties for traditional RNNs~\cite{radford_language_nodate, vaswani_attention_2017}. These properties make transformers a promising approach for predicting future human agent behaviour from historical patterns, hence we utilise them in our work.

\subsubsection*{Human Modelling}
A common drawback of deep learning models (such as MLPs, RNNs, transformers and RL methods) is that significant amounts of data are required for training \cite{radford_language_nodate, nguyen_review_2019}. One approach for coping with this issue is to develop a model that can simulate human behaviours. These models 
can include programming rules, imitation learning, and probabilistic models with or without Boltzmann rationality~\cite{yang_optimal_2022,reddy_where_2019,laidlaw_boltzmann_2022, chan_human_2021}. In the current work, 
our human agent model includes different systematic biases that allow us to simulate human agents with varying behaviours.

\section{PROBLEM DESCRIPTION}



The domain of the current work is human-robot agent cooperation where the human agent is included in the environment. Our focus is to obtain robot agent policies, using RL, that are able to cooperate with human agents that change behaviour over time, \textit{i.e.} solve a non-stationarity RL problem.

Formally, our approach follows the Meta-RL ("learning to learn") paradigm~\cite{bing_meta-reinforcement_2022} where the non-stationarity problem can be described as a zero-shot meta learning problem. This approach represents each behaviour with a latent state and treats them as separate tasks, while the robot agent adapts instantly within each task (\textit{i.e.} behaviour) ~\cite{bing_meta-reinforcement_2022, arango_multimodal_2021}.  
The Meta-RL setting that is relevant to the non-stationary issue requires 'zero-shot' adaptation where the agent may encounter a new task (\textit{i.e.} behaviour) at every timestep and react immediately \cite{bing_meta-reinforcement_2022}. The non-stationarity challenge faced by a robot agent when cooperating with a human agent is formalised below as zero-shot meta-learning.




\subsubsection*{Single-agent RL}
The environmental dynamics relate to the probability of reaching the next state $s'$ and the corresponding reward $r$, from the current state $s$ and robot action $a$. In single-agent RL, the environmental dynamics $P(s', r | s, a )$ comprise the reward $P(r | s, a )$ and the state transition dynamics $P(s' | s, a )$, and are constant over time.

\subsubsection*{RL with a human agent}
We consider a human agent behaviour model that depends on time, denoted as $a_H \sim \pi_H (t)$. In order to consider the human agent behaviour within the policy of the robot agent, we include the human agent in the environment, hence the environmental dynamics (transition function) are $P(s',r| s, a, a_H)$. Given that the human behaviour changes (i.e. $P(a_H | t)$ is a function of time), the environmental dynamics $P(s', r | s, a, t)$ are also a function of time, hence "non-stationary". 


\subsubsection*{Human agent behaviour} To model a time-dependent human agent we assume that the human actions are governed by a probability distribution conditioned on a history $h_{H,L}$. This is written formally as: 
\vspace{-2mm}
\begin{equation}\label{eq: action_sample}
a_H \sim \pi_H(a_H | h_{H,L}),
\vspace{-2mm}
\end{equation}
where 
\vspace{-2mm}
\begin{equation}\label{eq: human_history}
h_{H,L} = \{s_{t-L:t}, a_{t-L:t-1}, a_{H,t-L:t-1}\},
\end{equation}
and $s$ are states, $a_H$ are human actions, $a$ are robot actions and subscript $L$ indicates a window of L last timesteps. 
The human behaviour is assumed to be Markovian given the look-back window, $L$, hence  
\begin{align}
P(a_H=a_{H,t} | s_{L},a_{L},s_{t+1}) &= 1,
\end{align}
and this implies that  $a_{H,t}$ is known, given $\{s_{L}, a_t, s_{t+1}\}$.
This means that  the human action can be inferred from a lookback window of states $s_L$ and robot actions $a_{L}$, formally written as: 
\begin{equation}\label{eq: deterministic_simplify}
P(a_{H,t} | s_{L}, a_{L}).
\end{equation}






\subsubsection*{Optimal robot agent policy}

\begin{figure}[H]
\centering
\includegraphics[width=0.35\paperwidth]{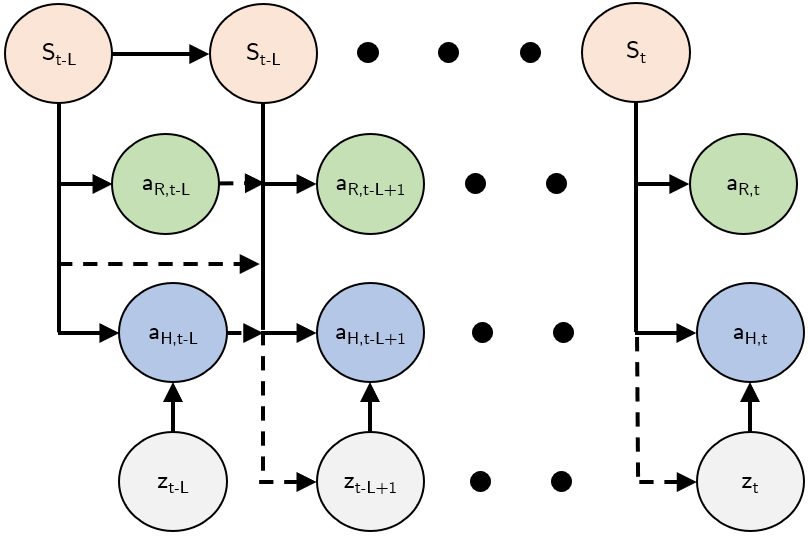}
\caption{Directed Acyclic Graph of the transition model.}
\label{fig:transition model}
\end{figure}

The robot agent takes a series of actions and observes a sequence of states and rewards. This allows it to obtain a trace up to time $t$ denoted as $h_R = \{s_{0:t}, a_{0:t-1}, r_{0:t-1}\}~\refstepcounter{equation}(\theequation)\label{eq: robot_history}$.
The aim of the robot agent is to maximise the total expected reward of the policy $r(s_t,a_t,a_H)$ over $N$ interactions: 
\begin{equation}
\label{eq:RL_objective}
\theta^* = \arg \max_{\theta} \sum^N_j \mathbb{E}_{\pi_{\theta} (\tau \sim p(\tau_t))} [\sum_{t=0}^T r(s_t, a_t, a_{H,t})],
\end{equation}
where $\theta^*$ are the optimal parameters of the robot agent policy and $p(\tau_t)$ is the probability for a trajectory of length $t$ given by: 
\begin{multline} \label{eq:2}
    p(\tau_t) = \rho(s_0) \prod_{t=0}^T \pi(a_t | h_{R,t-L:t}) \pi_H(a_{H,t} | h_{H,t-L:t}) \\ T(s_{t+1} | s_t, a_t, a_{H,t}),
\end{multline}
where $\rho_0$ is the initial state distribution, $\pi$ is the robot's policy, $\pi_H$ is the human's policy and $T$ is the environmental dynamics. In Figure~\ref{fig:transition model} we provide the directed acyclic graph (DAG) that captures the overall system and illustrates the transition model.

The proposed framework can be beneficial in instances of human-robot collaboration in tasks such as cooking. The state $s$ could include the user's location in the kitchen, the ingredients currently in use, the user's movements etc. The actions $a$ could range from handing over utensils, to stirring a mixture, to observing without intervening. The robot should be trained to interpret the user's needs. For instance, if the human is cutting vegetables, will they require a bowl for the cut pieces? If they are about to serve a piece of lasagna will they need a spatula? This need can be captured in the history of length $L$ ($h_{H,t-L:t}$ \eqref{eq: 2}) and from this the robot can predict whether to offer a bowl or spatula and provide it at the right moment. The reward $r$ can include the efficacy of the robot's actions in relation to the human's needs, and its ability to match the human's pace and avoid unnecessary interruptions. Overall, this framework can enable the robot to predict human needs in a cooking environment and thereby enhance the cooking process.

 

\subsection*{Computational formulation}
\label{subsec:comp_form}

The problem described above is formulated as learning an optimal policy in Markov Decision Process environments where the changes are a function of the robot’s policy. The tasks are modelled as a Markov Decision Process (MDP), $M_V=(S,A,R_v,P_v,\rho_0,\gamma)$, where $s \in S$ is the state space observed by the robot, $a \in A$ is the robot’s action space, $R_v : S \times A \times S \rightarrow R$ is the reward function, $T_v : S \times A \rightarrow \Delta(S)$ is the state transition probability function, $\rho_0$ is the initial state distribution, and $\gamma \in [0,1]$ is the discount factor. 

Following the meta-RL paradigm described above, each task (behaviour) is assumed to be represented by a latent vector (with parameters $v \in V \subset R^d$) that is Markovian and captures the changes in behaviour. This makes the environment dynamics a function of $v$, $P(s', r | s, a, v)$. The class of MDPs with this parameterisation is defined by $\mathcal{M} : = \{M_v : v \in V\}$. 

The robot agent selects its action $a_t \sim \pi_\theta (a | s_t, \hat{v_t})$, receives reward $r$, observes next state $s'$, and aims to maximise the expected cumulative reward defined in~\eqref{eq:RL_objective}.







\section{THE PROPOSED APPROACH}





We assume that the human behaviour is governed by a latent state $z$. Under this assumption, the human policy depends on the latent state $z$ and the current physical state $s$ (see Figure~\ref{fig:transition model}),  written with the conditional: $\pi_H (a_H | s, z)$. On this basis, the transition dynamics can be re-written to be conditioned on the latent state:
\begin{multline}\label{eq: deterministic1}
    p(s',r | s, a_R, a_H) = p(s', r | s, a_R, a_H, z) \\ \text{if} \hspace{0.2cm} s',r \perp\perp z | s, a, a_H
\end{multline}
\begin{multline}\label{eq: determinstic2}
    \implies p(s',r | s, a_R, a_H, z) = p(s', r | s, a_R, z) \\ \text{if} \hspace{0.2cm} s',r \perp\perp a_H | s, a_R, z.
\end{multline}
This makes the environmental dynamics a function of the human's latent state $z$. RL methods assume that the environment dynamics are Markovian, but the variable $z$ may not be Markovian. Hence, the latent state $v$ (see~\cref{subsec:comp_form}) is learned in a Markovian manner so that these methods can be applied successfully, converge to optimal policies, and capture the different behaviours. This approach provides time-invariant state transition dynamics $P(s', v', r | s, a, v)$, and allows the definition of a new state $s_v = [s, v]$ that reduces the transition dynamics to the single-agent MDP problem $P(s_v', r | s_v, a)$. This enables the application of classical RL techniques. In other words,
the problem  reduces to solving a single-agent RL problem, whilst inferring a Markovian latent state representation $\hat{v_t}$, where: 
\begin{equation} \label{eq:infer latent}
    \hat{v_t} = f(z_{t-L}:z_t) = f(h_{R,L})
\end{equation}
and $\hat{v_t}$ is a function of the robot's history, that is:
\begin{equation} \label{eq:robot_history_L}
    h_{R,L} = \{s_{t-L:t}, a_{t-L:t-1}, r_{t-L:t-1}\}.
\end{equation}



\subsection{Human agent model}\label{human_model_section}

We developed a  probabilistic human agent model with  attributes that encapsulate the key challenges that arise when interacting with different humans whose behaviour, $\pi (a_H | s, t)$, changes over time. 
The three key attributes of our human agent model are: (i) a time-invariant component; (ii) a low-frequency history-dependent (hysteresis) component; (iii) a high-frequency history-dependent component. Each attribute had a corresponding underlying population-level distribution weight $w$. Each simulated human agent was created by sampling a weight from these three distributions $(w_{c}, w_{lf}, w_{hf})$. These weights define how the human agent behaviour evolves as a function of its experience over time. 
The ground truth latent state evolves as a function of the history-dependent human dynamics, as described below:
\begin{equation}
z_t \sim \rho_z(z_t | s_{t-L : t}, a_{t-L : t} ; w_{human}), 
\end{equation}
where 
\begin{equation}\label{eq: weight}
    w_{human} \in (w_{c}, w_{lf}, w_{hf})
\end{equation}
The time-invariant latent factor ($z_{c}$) varies between different humans agents but is constant over time for a given human. The time-dependent factors ($z_{lf}, z_{hf}$) evolve over time. The low-frequency component ($z_{lf}$) depends on the previous episode, and the high-frequency component ($z_{hf}$) depends on the previous action. The three components of the human behaviour are combined together with different weights according to: 
\begin{equation} \label{eq: 2}
\Tilde{p}(z) = \sum^N_i w_i p_i (z) = w_c z_c + w_{lf} z_{lf} + w_{hf} z_{hf}
\end{equation}
to simulate different humans agents. The human agent behaviour is defined so there is one-to-one mapping between the ground truth latent state and where the human agent hits the ball. This is written formally as: 
\begin{equation}
a_{H,0} \sim P(a_{H,0} | z) = \frac{i}{\sum_i z_i} [z_1, z_2, z_3, z_4]
\end{equation}
where each latent component $z_i$ represents a different goal location.

This model simulates human behaviour to generate human-like data. Existing research indicates that human behaviour is influenced by time-invariant elements plus components dependent on varying frequencies (such as low and high frequency factors) outlined in the human model (\ref{human_model_section})  \cite{ditterich_evidence_2006}. However, this simplified model may not capture the contextual relationships inherent in human behaviour. 

The weight terms in \eqref{eq: 2}, \eqref{eq: weight} determine to which extent each component affects the human's behaviour. For example, a low-frequency weight $w_{lf}$ set to 0 indicates that the low-frequency component has no affect on the decisions made by the human model.






\subsection{BeTrans}
 \begin{figure}[t]
\centering
\includegraphics[width=0.38\paperwidth]{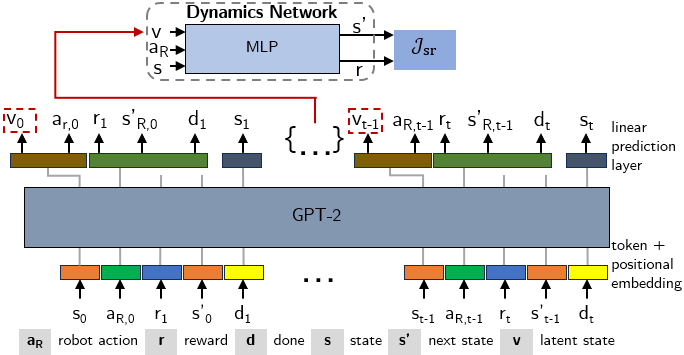}
\caption{Schematic illustrating the proposed BeTrans approach.}
\label{fig:BeTrans}
\end{figure}


We utilised an auto-regressive transformer architecture to learn a Markovian latent state representation $\hat{v_t}$ that captures the changes in the human agent behaviour (see \eqref{eq:infer latent}). The transformer model is based on GPT-2 \cite{radford_language_nodate}. This includes a causal attention mask to ensure predictions are based only on previous (in time) data. We used the smallest GPT-2 model with 12 multi-attention heads \cite{radford_language_nodate} as it is sufficient for this problem and can perform faster inference - though larger GPT-2 models provide greater accuracy. 
The self-attention mechanism is used to learn various contextual nuances of the behaviour. 
We found that for our relatively simple environment this model gave adequate precision.

Figure \ref{fig:BeTrans} illustrates theapproach. At each time step a tuple ($s_t,a_t,r_t,s_{t+1},d_{t})$ is passed into the architecture as separate tokens so the experiences form a long trajectory of states $s_t$, actions $a_t$, rewards $r_t$, next states $s_{t+1}$ and done tokens $d_t$ (across many episodes). $\tau = (s_0, a_0, r_1, s_1, d_1, s_1, ... , s_{t-1}, a_{t-1}, r_t, s_t, d_t)$ where $t$ is the number of time steps. The states are normalised between -1 and 1 and the actions are one-hot encoded. Each token is embedded into a $H=192$ dimensional vector. A separate linear layer is used to embed each token type (i.e. $s,a,r,d$) in its corresponding $H$ dimensional vector. The model utilises the tokens up to the state to predict the latent state $v$ and subsequent action $a$. It uses all tokens up to and including the action to predict the reward, next state, and whether the task is finished ('done') $(r, s', d)$ (see Figure \ref{fig:BeTrans}). 
 In a similar manner to the GPT-2 model, absolute positional embeddings are added \cite{radford_language_nodate}.

The transformer was trained in two phases - a pre-training phase and a fine-tuning phase. In the pre-training phase, the losses from predicting $(s,a,r,s',d)$ were used to update the transformer's weights $\theta_{BT}$. The purpose of this phase was to train the transformer to learn the underlying structure of the data. The data used to generate the replay buffer for pre-training were generated by taking random actions for 10,000 time steps. In the fine-tuning phase, the predicted latent variable $v$ is fed into the Dynamics Network (DN) along with the corresponding state(s) $s$ and action(s) $a$ at that time step to predict the next state(s) $s'$ and reward(s) $r$. The DN is a multi-layer perceptron (MLP) and consists of two output heads for $s'$ and $r$. The DN consists of two-hidden layers of size 256. In the case where $v$ is Markovian for each timestep, the DN takes in $(s_t, v_t)$ and predicts $(s_{t+1}, r_{t+1})$. In the case where $v$ is non-Markovian given the previous timestep, the transformer predicts a forecast horizon of states and rewards $(s_{t+1:t+N}$, $r_{t+1:t+N})$, where $N$ is the horizon window. We assume that the human agent's behaviour is constant during a horizon window. The reconstruction loss is $\mathcal{J}_{sr} = \mathbb{E}[(s - \hat{s})^2 + (r - \hat{r})^2]$.

We investigated both continuous and discrete latent representations. For discrete latent variables, the Gumbel-Softmax distribution was used to enable backpropagation \cite{jang_categorical_2017}. For continuous latent variables, the Kullback-Leibler (KL) divergence was used  to reduce overfitting~\cite{jiang_transformer_2020}. This was important as the flexibility provided by the transformer's architecture could inadvertently incorporate extraneous noise.  The KL divergence was taken between the approximate posterior distribution and prior distribution $p(v) = \mathcal{N}(0,\mathcal{I})$. The posterior distribution was a diagonal Gaussian where the mean and standard deviation came from the transformer's two latent variable output heads (the transformer has just one output head for the case where the latent variable is discrete). The loss $\mathcal{J}_{DN} = \mathcal{J}_{sr} + \mathcal{J}_{KL}$ was used to update the transformer $\phi_{BT}$ and Dynamics Network's weights $\theta_{DN}$.

During training, data were sampled from the replay buffer (which stores the experiences) with a batch size of 32. A block size $L$ of 125 was chosen for the transformer so that it could look-back 25 timesteps (see (\ref{eq:infer latent})). Different block sizes could be chosen depending on the desired look-back length. The proposed approach leverages the transformer's employment of parallelism, positional encoding, and self-attention mechanisms to effectively capture historical dependencies to learn the underlying latent variable.




\subsection{RL method}
\subsubsection*{RL with latent strategies}
For the RL element of our approach, we used the soft-actor critic (SAC) algorithm \cite{haarnoja_soft_2018}, which is an off-policy algorithm conditioned on the latent state. The actor $\pi$ with parameters $\theta_\pi$ and critic $Q$ network with parameters $\theta_Q$ were trained with the SAC loss $\mathcal{J}_{\pi}$ and $\mathcal{J}_{Q}$ \cite{haarnoja_soft_2018}. The hyperparameters in this base algorithm were taken from CleanRL \cite{huang_cleanrl_nodate}. It was found that the default hyperparameters were sufficient for the method to learn an effective policy with an appropriately tuned reward function. In SAC, experiences (i.e. $s,a,r,s',d)$) are sampled at random from the replay buffer. The transformer infers from the history window of desired length (in this case the last 25 timesteps) the corresponding latent variable $v$ at a given timestep. During training, every 1000 timesteps, five batches are sampled from the replay buffer, the SAC ($\mathcal{J}_{\pi}$, $\mathcal{J}_{Q}$) and DN $\mathcal{J}_{sr}$ losses are calculated, and the transformer and SAC weights are updated - see Algorithm~\ref{alg:BeTrans_algorithm}. The learning rate for both the policy $\alpha_\pi$ and critic networks $\alpha_Q$ was set to $3 \times 10^{-4}$. The learning rate for the behavioural transformer $\alpha_{BT}$ and DN $\alpha_{DN}$ was $4 \times 10^{-4}$ and $1 \times 10^{-3}$ respectively.





\begin{algorithm}[t]
\caption{Behavioural Transformer}\label{alg:BeTrans_algorithm}
\begin{algorithmic}
\Require $\text{Learning rates } \alpha_Q, \alpha_\pi, \alpha_{\text{BT}}, \alpha_{\text{DN}}$
\State $\text{Randomly initialise } \theta_Q, \theta_\pi, \text{ and } \theta_{DN}$
\State $\text{Initialise $\phi_{BT}$} \text{ from pre-training}$
\State $\text{Initialise empty replay buffer } \beta$
\For{$\text{episode} = 1, 2, ... $} \Comment{Collect data}
\For{$t = 1, 2, ... $}
\State $\hat{z_t} \gets f(h_R)$
\State $a_t \sim \pi_{\theta} (a | s, \hat{z_t})$
\EndFor
\State $\text{Assign } \beta[i] \gets \tau^i$
\If{$\text{update }$}\algorithmiccomment{Update GPT and SAC}
\For{$\text{update in } 1, 2, ...$}
\State{sample from batch $\beta$}
\State $\phi_{BT} \gets \phi_{BT} - \alpha_{BT} \nabla_{\theta_{BT}} (\mathcal{J}_{sr} + \mathcal{J}_{KL})$
\State $\psi_{DN} \gets \psi_{DN} - \alpha_{DN} \nabla_{\theta_{DN}} \mathcal{J}_{sr}$
\State $\theta_Q \gets \theta_Q - \alpha_Q \nabla_{\theta_Q} \mathcal{J_Q} $
\State $\theta_\pi \gets \theta_\pi - \alpha_\pi \nabla_{\theta_Q} \mathcal{J_\pi}$
\EndFor
\EndIf
\EndFor

\end{algorithmic}
\end{algorithm}

\section{RESULTS}

We performed experiments with a range of simulated human agents and customised environment parameters. We tested the efficacy of our approach by comparing our algorithm against four baselines over a range of parameters within the custom environment. The human agent behaviour was constant during an episode (always passing the ball to one of the target locations) but changed between episodes. The environmental conditions that we considered were; (i) introducing noise in the observations of the ball position; (ii) changing human behaviour during an episode; and (iii) selecting the human's goals based on a function of longer history.

We also performed experiments across a range of environmental parameters, which are outlined in section \ref{simulation_platform}. We investigated the impact of learning a continuous or discrete latent variable on the performance of the proposed method.

\subsection{Simulation Platform}
There are a number of multi-agent environments~\cite{papoudakis_benchmarking_2021} available, but we created our own custom environment ("co-pass") that allowed us to have complete parametric control over factors such as:

\begin{itemize}
    \item \textit{Reward sparsity}: number of timesteps it takes the ball to travel between paddles
    \item \textit{Task difficulty}: ratio of the number of locations where the human can pass the ball against the time it takes the ball to travel between paddles
    \item \textit{Observability}: amount of noise in the observations of the ball positions in the environment, and whether the ball is observed 
    \item \textit{Simulated human}: The human policy as a function of its historical dependencies
\end{itemize}
Co-pass provided a test bed for the algorithms and a platform for investigating our hypotheses (our primary hypothesis being that BeTrans would effectively model historical dependencies). This allowed us to evaluate whether the robot agent could adapt to changing human behaviour during an episode in the presence of observation noise.

\begin{figure}
\centering
\includegraphics[width=0.4\paperwidth]{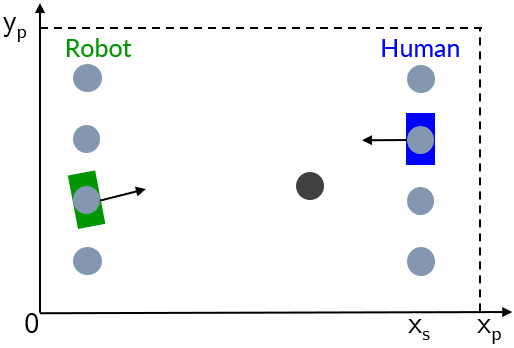}
\caption{Schematic of the custom environment co-pass.}
\label{fig:game schematic}
\end{figure}

The experimental design was a simulated collaborative 'ball passing' task between a robot and human agent (as shown in Fig. \ref{fig:game schematic}).  
We discretised the state space and varied both the number of target locations and the timesteps over which the ball travelled between the paddles. The dynamics were deterministic with the ball changing direction when it hit a paddle and the subsequent trajectory being a function of the paddle orientation. We sampled the position of the ball from a standard normal distribution (with the mean at its 'true' position) when injecting noise into the observations of the ball's position. A point was awarded for each successful intercept of the ball. The goal was to accumulate as many points as possible within a set time frame (equivalent to the tennis rally example provided in the introduction).


\label{simulation_platform}

\subsection{Experiment}

We compared our approach against the hypothetical ideal method and three baselines: 
\begin{itemize}
    \item \textbf{Oracle:} This approach 'knows' the ground truth latent state of the human and provides an upper bound on the performance of a method 
    \item \textbf{SAC:} This approach does not explicitly model the dynamics of a human \cite{haarnoja_soft_2018-1}
    \item \textbf{LILI:} This approach uses a variational autoencoder to learn the dynamics of a human and provides a baseline comparison \cite{xie_learning_2020}
    \item \textbf{RNN-LILI:} This method is analogous to LILI, but employs a RNN as the encoder block, similar to \cite{parekh_rili_2022}
    \item \textbf{BeTrans:} Our approach - using a conditional transformer to learn the dynamics of a human
\end{itemize}

In the first experiment, the episodes had a fixed length and the human agent latent state (goal) - which determined where the human agent would pass the ball to - was constant during episodes and only changed between episodes. In the second experiment, noise was added to observations of the ball position, by sampling from a normal distribution with a variance of 1 unit in the x-position and 0.5 units in the y-position of the ball, where 1 unit represents the x-distance traveled by the ball between the paddles in one timestep and the y-distance between two consecutive target locations, respectively. For the third experiment, the latent state governing the human behaviour changed during episodes (thus changing where the human agent would pass the ball). In the fourth experiment, longer history lengths were included to investigate BeTrans's ability to cope with long-term dependencies. 

For these experiments, the robot agent was trained with simulated human agents whose weights $(w_{c}, w_{lf}, w_{hf})$ were sampled from a Continuous Bernoulli distribution~\cite{loaiza-ganem_continuous_2019}. These determined how the human behaviour evolved, as detailed in~\cref{human_model_section}.

Co-pass's parameters were changed to investigate the effect they had on performance. These changes included reward sparsity, defined as the number of steps between the paddles (varied between 1 and 20), and the number of target positions where the human could pass the ball (varied between 3 and 40). The ratio between the number of targets and steps between the paddles (task difficulty) was kept greater than 2 to ensure that predictions about the human agent  were relevant for successful collaboration.



\begin{figure*}[h]
\centering
  \includegraphics[width=\linewidth]{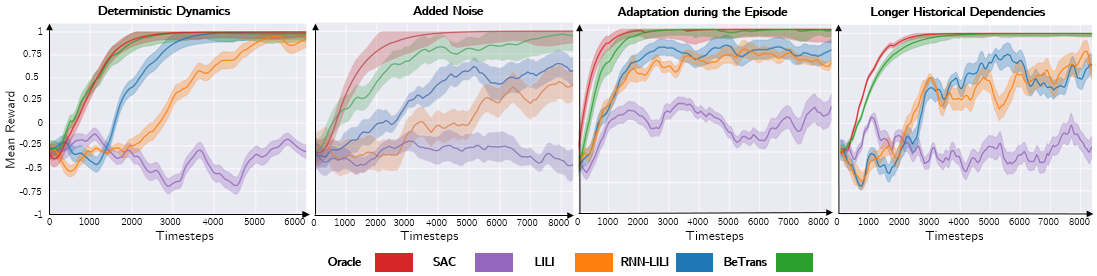}
  \caption{A schematic comparing the BeTrans (our) approach against the baseline methods, evaluated on the custom environment. The lines represent the average of 10 random seeds, evaluated at held out testing episodes. The shaded ribbons illustrate the standard error for these averages.}
  \label{fig:four_plots_fig}
\end{figure*}

\subsection{Analysis}

Co-pass was used to evaluate the efficacy of the methods, with 10 random seeds being employed for each experiment. The findings from four of the experiments are presented in Figure \ref{fig:four_plots_fig}. Notably, the proposed approach outperforms the state-of-the-art algorithms across the experiment, and approaches the oracle's level of performance. In these experiments, it is possible for the robot agent to reach the maximum reward as the human agent's behaviour is deterministic given the interaction history, so a perfect prediction of the human actions is possible.

\textbf{Fixed behaviour within episodes.} In the first experiment, the human agent's latent state changed between episodes but not during an episode. Figure \ref{fig:four_plots_fig} (first panel) shows all approaches other than SAC (which does not model the dynamics of the human agent) converged to the optimal policy. Notably, BeTrans showed the fastest convergence.

\textbf{Adding noise to the observations.} In the second experiment, Figure \ref{fig:four_plots_fig} (second panel) shows that BeTrans is robust in modelling behavioural dynamics even in the presence of significant levels of noise. 

\textbf{Changing behaviour within episodes.} In the third experiment, we evaluated our approach in scenarios where the latent state governing the simulated human behaviour changed during episodes (third panel of Figure \ref{fig:four_plots_fig}). In this case, BeTrans leverages its flexibilty to converge to an optimal policy. This contrasts with LILI and RNN-LILI, which assume constant human behaviour during episodes, and SAC (which does not model human dynamics).

\textbf{Longer historical-dependencies.} For the fourth experiment (rightmost panel of Figure \ref{fig:four_plots_fig}), the human agent's behaviour depended on a longer history of interaction (the previous 10 episodes). For these longer historical dependencies, there was a greater discrepancy between the performance of the BeTrans and the baseline methods. We postulate that this is due to the transformer's superior performance in capturing long-term historical-dependencies, due to its utilisation of positional encodings. This enables the model to effectively process long-range dependencies compared to the MLP and RNN-based approaches.

\textbf{Changing environmental parameters}. It took BeTrans 31\% longer to converge to an optimal policy when the number of timesteps, corresponding to reward sparsity, doubled from 5 to 10.  All methods converged to the optimal policy when the task difficulty ratio was less than 2 as predictions of the human behaviour were not required.

\textbf{Continuous and discrete latent variables}. We investigated scenarios where the transformer output both discrete and continuous latent variables. A gumbel-softmax was used in the discrete case. The RL policy learned 24-46\% faster with discrete than continuous variables for the custom environment without noise. This implies that noise was picked up in the latent variable representation when using the smooth distance metric in the continuous case. Thus, discretising the latent variable can act as a regulariser. In the discrete case, the method required the output dimension to be greater than or equal to the number of possible human goals. This corresponds to the number of target locations where the human agent can pass the ball in this case.

\begin{figure*}[t]
    \centering
    \subfloat[First snapshot]{\includegraphics[width=0.225\linewidth]{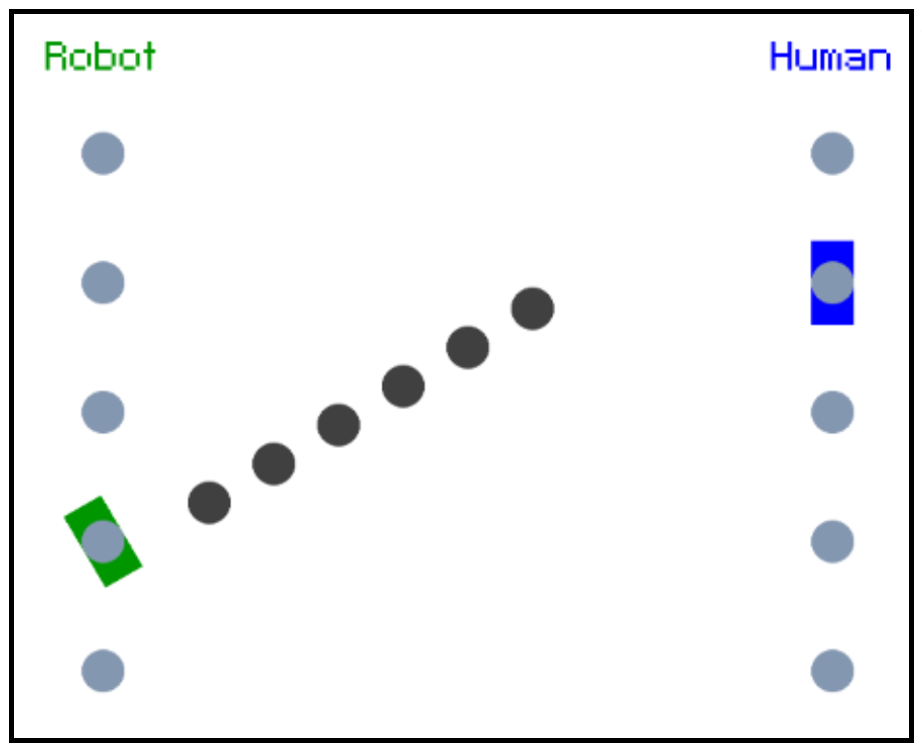}}~ 
    \subfloat[Second snapshot]{\includegraphics[width=0.225\linewidth]{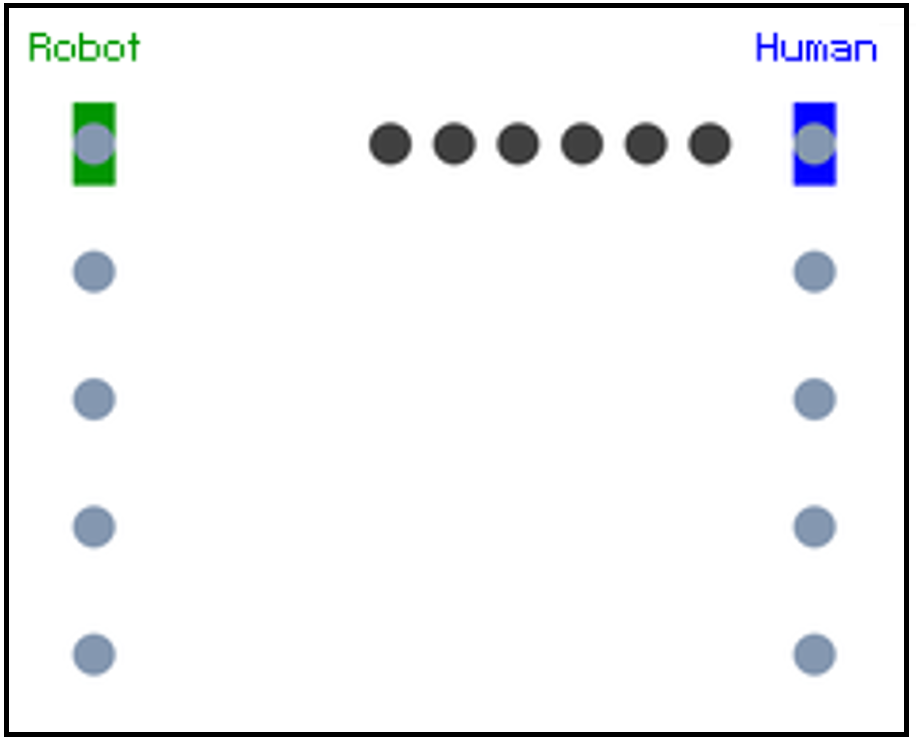}}~
    \subfloat[Third snapshot]{\includegraphics[width=0.225\linewidth]{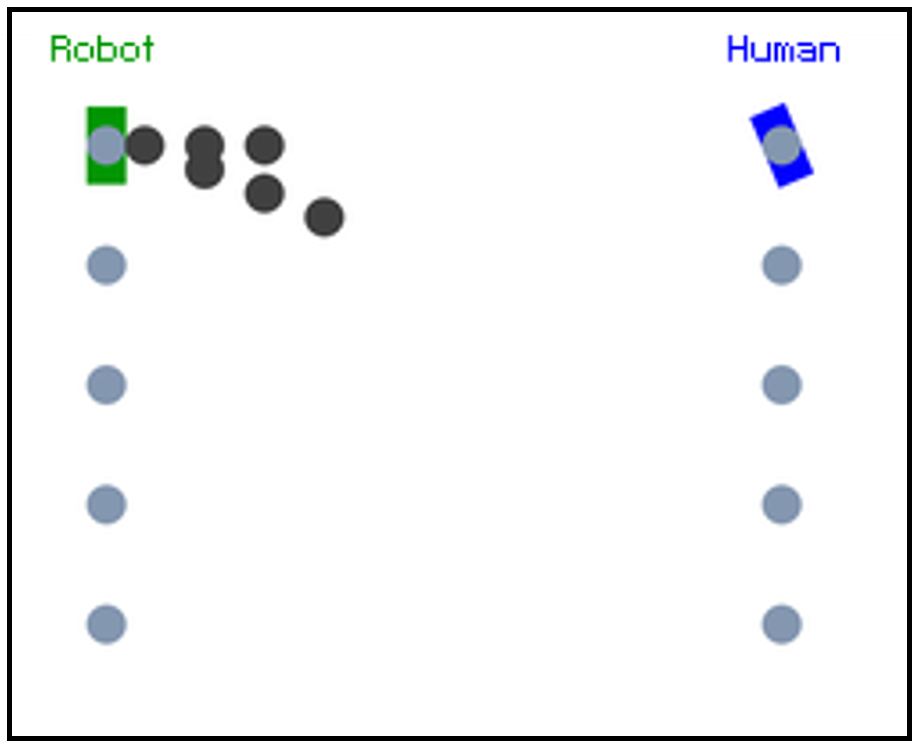}}~
    \subfloat[Fourth snapshot]{\includegraphics[width=0.225\linewidth]{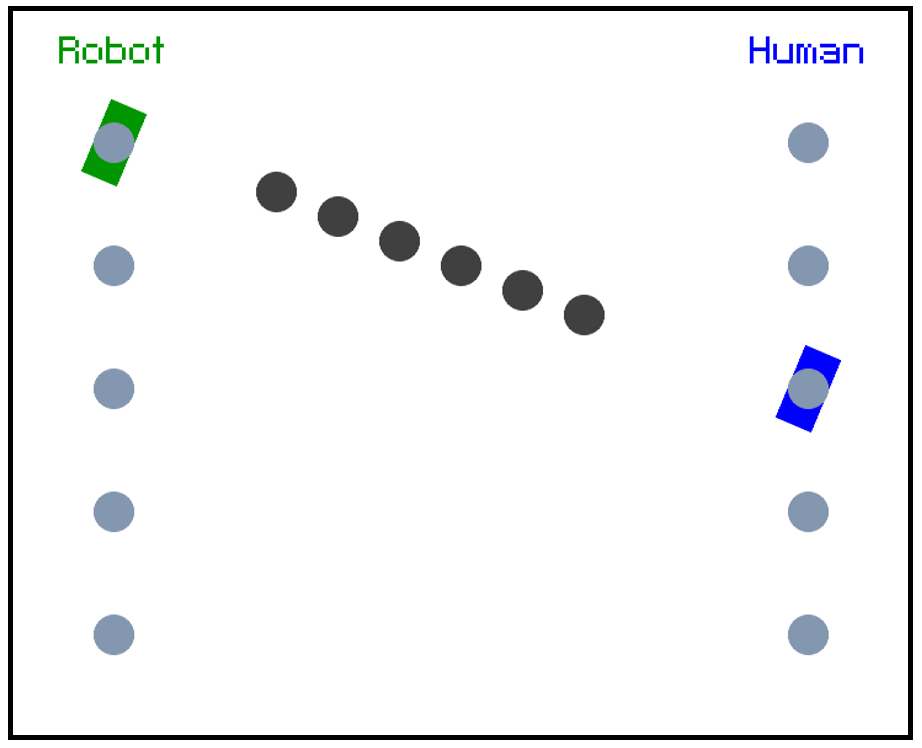}}~
    \caption{Several snapshots of the ball's positions in co-pass when the ball is moving from the robot agent to the human agent's paddle.} 
    \label{fig:snapshots}
\end{figure*}

\begin{figure}[t]
    \centering
    \subfloat[Untrained policy]{\includegraphics[width=0.45\linewidth]{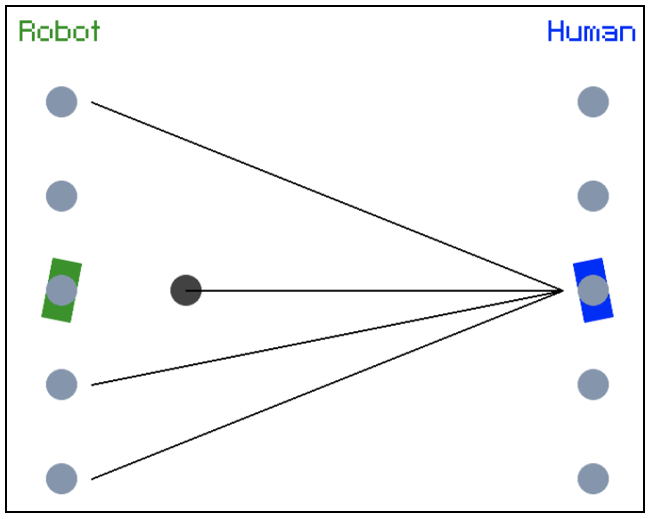}}~ 
    \subfloat[Trained poilcy]{\includegraphics[width=0.45\linewidth]{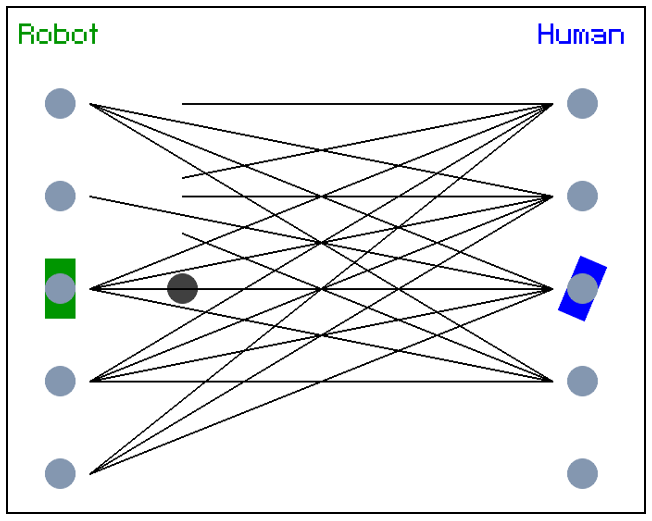}}~
    \caption{Traces of the ball when the robot's policy is evaluated over five episodes. The maximum number of hits in each episode is five. Traces are only kept if the ball hits the robot agent's paddle.} 
    \label{fig:traces}
\end{figure}

An illustration of 'co-pass' in action is depicted in Figure \ref{fig:snapshots} where several snapshots of the ball travelling between the two paddles are shown. Figure \ref{fig:traces} shows the traces when evaluating an untrained and a trained robot agent policy with a human agent over five episodes, where the maximum number of passes in each episode is five. The trained robot agent is able to predict where the ball will go and intercepts it more frequently. The greater number of lines show that more passes took place with the trained robot agent.

\section{CONCLUSIONS}
We have created a framework that can learn optimal robot policies in human-robot agent collaboration scenarios where the human agent changes behaviour. We validated our method on a novel environment with customisable parameters and over a range of simulated human agents. The approach was trained against simulated human agents with systematic biases, but whose behaviour is Markovian given sufficient history. Our novel BeTrans methodology allows the robot agent to adapt to changing human behaviours.

We provide a theoretical framework to show that our approach allows the robot to learn a representation which makes the environmental dynamics time-invariant, thereby allowing the application of single-agent RL methods. The human behaviour is assumed to be governed by a latent state, and the method adapts to state transitions during episodes and can handle noise. BeTrans improves on existing state-of-the-art latent variable RL methods.

\textbf{Limitations and future work:} Our method allows the robot to effectively learn a latent variable and interact with humans whose behaviour can change at any timestep - but assumes that the behaviour is Markovian and the dynamics are deterministic. The assumption that the latent state is Markovian represents a simplification of the problem. Single-agent RL methods are based on a Markov Decision Process that assumes each state has sufficient information to predict the probability distribution over the next states given an action. Thus, if the latent state (human's goal) is non-Markovian, the policy cannot predict the next goal so it becomes a multi-task problem. In a multi-task problem, the robot can learn an optimal policy that maximises the reward for a given goal, but it cannot predict how its actions will affect goal transitions. If the robot cannot make this prediction then it is unable to influence the human's goal (as the human's goal transitions may be a function of the robot's actions). 

In future work, we plan to extend our approach to deal with non-Markovian behaviours and non-determinstic dynamics. We also plan to train our method using actual human data while performing a ball passing task. Furthermore, in the next phase of work we will 
validate our method with actual humans interacting with robots.






\end{document}